\documentclass[10pt, a4paper]{article}
\usepackage{lrec}
\usepackage{multibib}
\newcites{languageresource}{Language Resources}
\usepackage{graphicx}
\usepackage{tabularx}
\usepackage{soul}
\usepackage{multirow}
\usepackage{color}

\usepackage{epstopdf}
\usepackage[latin1]{inputenc}

\usepackage{hyperref}
\usepackage{xstring}

\title{eSCAPE: \\ a Large-scale \textit{S}ynthetic \textit{C}orpus for \textit{A}utomatic \textit{P}ost-\textit{E}diting}

\name{Matteo Negri$^{(1)}$, Marco Turchi$^{(1)}$, Rajen Chatterjee$^{(1,2)}$, Nicola Bertoldi$^{(1)}$}

\address{$^{(1)}$Fondazione Bruno Kessler, $^{(2)}$University of Trento \\
         Trento, Italy \\
         \{negri, turchi, chatterjee, bertoldi\}@fbk.eu\\}

\abstract{
Training models for the automatic correction of machine-translated text usually relies on data consisting of (\textit{source}, \textit{MT}, \textit{human\_post-edit}) triplets providing, for each source sentence, examples of translation errors with the corresponding 
corrections made by a human post-editor. 
Ideally, a large amount of data of this kind should allow the model to learn reliable correction patterns and effectively apply them at test stage on unseen (\textit{source}, \textit{MT}) pairs. In practice, however, their limited availability calls for solutions that also integrate in the training process other sources of knowledge. Along this direction, state-of-the-art results have been recently achieved by systems that, in addition to a limited amount of available training data, exploit  artificial corpora that approximate elements of the ``gold'' training instances with automatic translations.
Following this idea, we present eSCAPE, the largest   freely-available \textit{S}ynthetic \textit{C}orpus for \textit{A}utomatic \textit{P}ost-\textit{E}diting released so far. eSCAPE consists of millions of entries in which  the \textit{MT} element of the training triplets has been obtained by translating the \textit{source} side of publicly-available parallel corpora,  and using the \textit{target} side as an artificial human post-edit. Translations are obtained both with phrase-based and neural models. 
For each MT paradigm, eSCAPE contains  7.2 million triplets for English--German and 3.3 millions for English--Italian, resulting in a total of 14,4 and 6,6 million instances respectively. 
The usefulness of eSCAPE is proved through experiments in a general-domain scenario, the most challenging one for automatic post-editing. For both language directions, the  models trained on our artificial data always improve  MT quality with statistically significant gains. The current version of  eSCAPE can be freely downloaded from: \url{http://hltshare.fbk.eu/QT21/eSCAPE.html}.
\\ \newline  }

\begin{document}

\maketitleabstract

\section{Introduction}
Automatic post-editing (APE) for machine translation (MT) aims to fix recurrent errors made by the MT decoder by learning from correction examples.
As a post-processing step, APE has several possible applications, especially in \textit{black-box} scenarios (\textit{e.g.} when working with a third-party translation engine) in which the MT system is used ``as is" and is not directly   accessible for retraining or for more radical internal modifications. In such scenarios, as pointed out by \newcite{Chatterjee-EtAl:2015:acl}, APE systems can help to: \textit{i)} improve MT output by exploiting information unavailable to the decoder, or by performing a deeper text analysis that is too expensive at the decoding stage; \textit{ii)} provide professional translators with improved MT output quality to reduce (human) post-editing effort, and \textit{iii)} adapt the output of a general-purpose MT system to the lexicon/style requested in a specific application domain.

The training of APE systems usually relies on data sets comprising (\textit{source}, \textit{MT}, \textit{human\_post-edit}) triplets, in which the \textit{source} sentence in a given language has been automatically translated to produce the \textit{MT} element that, in turn, has been manually corrected to produce the \textit{human\_post-edit}. In this supervised learning setting, the goal is to learn from the training data  (and possibly generalise) the appropriate corrections of systematic errors made by the MT system, and apply them at test stage on unseen (\textit{source}, \textit{MT}) pairs.
BLEU \cite{Papineni:2002} and TER \cite{snover2006study} computed against reference human post-edits are the standard evaluation metrics for the task.  Their respective improvements and reductions are usually compared against the baseline scores obtained by the original MT output that has been left untouched (\textit{i.e.} raw, non post-edited translations).

Early works on this problem date back to \cite{Allen:2000,simard2007statistical}, which addressed the problem as a ``monolingual translation'' task in which  raw MT output in the target language has to be translated, in the same language, into a fluent and adequate translation of the original source text. Although the general monolingual translation approach to the problem is still the same, over the years the proposed solutions evolved in several ways, first by refining the decoding approach and then, in the last couple of years, by radically changing the core APE technology. 

Decoding refinements successfully explored, for instance,  the integration of \textit{source} information for enhanced  (joint, context-aware)  input representation, either in the standard phrase-based MT (PBMT) framework \cite{Bechara2011} or in more elegant 
batch factored  models\cite{chatterjee-EtAl:2016:WMT} and online PBMT models \cite{chatterjee-EtAl:2017:EACLlong}. 
More recently, radical paradigm changes followed the ``neural revolution'' witnessed in the MT field.  The current state of the art is indeed represented by single/multi-source  neural APE systems, the former relying on the log-linear combination of monolingual and bilingual models \cite{junczysdowmunt-grundkiewicz:2016:WMT}, and the latter learning from source and target information in a joint fashion \cite{FBK:2017:WMT}. 
Recent works  addressed the problem  by also integrating external information such as word-level quality estimation scores \cite{chatterjee-EtAl:2017:WMT1} as a way to guide neural APE decoding towards better corrections.

Unsurprisingly, the impressive gains achieved by the neural solutions come at the cost of a much higher data demand compared to the PBMT methods.  To overcome this problem, the latest published results on neural APE have been obtained by exploiting synthetically-created data 
during training \cite{junczysdowmunt-grundkiewicz:2016:WMT,AMU:2017:WMT,CUNI:2017:WMT,DCU:2017:WMT,FBK:2017:WMT}.

These trends, which emerged after three rounds of the APE task  organised within the Conference on Machine Translation (WMT) \cite{bojar-EtAl:2015:WMT,bojar-EtAl:2016:WMT1,bojar-EtAl:2017:WMT1}, clearly indicate that: \textit{i)} information from the \textit{source} text is definitely useful to train reliable APE models, and  \textit{ii)}  the limited availability of ``gold'' training corpora made of (\textit{source}, \textit{MT}, \textit{human\_post-edit}) triplets calls for workarounds to unleash the full potential of state-of-the art but data-demanding neural systems. 

The eSCAPE corpus presented in this paper meets such demand by providing APE research with a large-scale synthetically-created data set consisting of millions of triplets for two language pairs: English--German and English--Italian. 
Starting from a collection of publicly-available parallel corpora, it was built by automatically translating the \textit{source} element of each sentence pair both with phrase-based an neural MT models, and using the original  \textit{MT} element as representative of a possible human correction.

This paper reports on the initial part of a roadmap aiming at more ambitious objectives. 
Future releases of the corpus will indeed include larger volumes of instances (translated with both MT paradigms) covering a larger spectrum of language combinations.
The following sections provide an overview of the existing resources 
(Section~2), a description of eSCAPE (Section~3) and a discussion of experiments with the corpus (Section~4).

\section{Related Work: Existing APE Corpora}
\label{sec:relworks}

\begin{table*}[ht]
\begin{center}
\begin{small}
\begin{tabular}{|l|c|c|c|c|c|}

\hline
\textbf{Corpus} & \textbf{Type} & \textbf{Languages} & \textbf{Domain} & \textbf{Size} & \textbf{Post-edits}\\
\hline
\cite{specia-etal_LREC:2010} & Gold & En-Es & LEGAL & 4K & Professional\\ \hline
\cite{Specia_EAMT:2011} & Gold & Fr-En/En-Es  & NEWS &  2.5K/1K  & Professional\\ \hline
\cite{zhechev:2012} & Gold & En-Ch/Cs/Ff/De/Hu/It/Ja/Ko/Pl/Br/Pt/Ru/Es & IT & 30K-410K & Professional\\ \hline
\cite{Potet:2012} & Gold & Fr-En & NEWS & 11K & Crowd\\ \hline
\cite{bojar-EtAl:2015:WMT} & Gold & En-Es & NEWS & 12K &  Crowd\\ \hline
\cite{bojar-EtAl:2016:WMT1} & Gold & En-De & IT & 13K &  Professional\\ \hline
\cite{bojar-EtAl:2017:WMT1} & Gold & En-De/De-En & IT/MEDICAL & 13K/26K &  Professional\\ \hline \hline
(Junczys-Dowmunt & Artificial & En-De & IT & 4.3M & -\\
and Grundkiewicz, 2016) &  &  &  &  & \\
\hline
\end{tabular}
\caption{Inventory of existing APE corpora}
\label{fig:inventory}
\end{small}
 \end{center}
\end{table*}

The growing interest towards APE has to confront with the hard truth of data scarcity. Although nowadays post-edited data are a clear by-product of industrial translation workflows, the largest part of the daily work done by professional translators focuses on proprietary or copyright data that cannot be released. Though present in the industrial sector (as confirmed by recent works coming from big players like SYSTRAN \cite{systran16} or eBay \cite{mathur:INLG17}),
APE technology is still more a matter of in-house development rather than a framework motivating free data sharing. 

The few existing  corpora that are usable for APE research can be classified into two types: \textit{i)} the aforementioned ``gold'' data sets made of (\textit{source}, \textit{MT}, \textit{human\_post-edit}) triplets, and \textit{ii)} the synthetic ones, to which our eSCAPE corpus belongs, in which some elements of the triplets derive from automatic translation. The remainder of this section provides an inventory of the existing APE corpora. As also shown in Table~\ref{fig:inventory}, the global picture is quite fragmented, with  domain-specific  data sets covering different language pairs, containing different types of post-edits and, most importantly, usually featuring a relatively small size.

\subsection{``Gold'' corpora}
The Autodesk Post-Editing Data corpus \cite{zhechev:2012}\footnote{\url{https://autodesk.app.box.com/Autodesk-PostEditing}} is one of such resources. It mainly covers the domain of software user manuals, with English sentences translated with  Autodesk's in-house PBMT system into several languages (simplified and Traditional Chinese, Czech, French, German, Hungarian, Italian, Japanese, Korean, Polish, Brazilian 
Portuguese, Russian, Spanish)  with between 30,000 and 410,000 segments per language. Post-edits are made by professional translators. 

Part  of the Autodesk  corpus has been used by \newcite{Chatterjee-EtAl:2015:acl} to compare different APE techniques in a controlled setting. For six target languages  (Czech, German, Spanish, French, Italian and Polish), this subset comprises around 16,000 (\textit{source}, \textit{MT}, \textit{human\_post-edit}) triplets that share the same English \textit{source}.
To ease the replicability of their experiments and the reuse  of the   selected  triplets, the authors released the scripts used for data extraction.\footnote{\url{https://bitbucket.org/turchmo/apeatfbk/src/master/papers/ACL2015/}} 

Another useful resource is described in \cite{Potet:2012}.\footnote{\url{http://www-clips.imag.fr/geod/User/marion.potet/index.php}} It consists of 10,881 triplets in which a French source sentence taken from several news corpora is translated into English by a PBMT system. Post-edits were collected using Amazon Mechanical Turk following strict control reviewing procedures to guarantee correction quality.

Two smaller corpora are respectively described in \cite{specia-etal_LREC:2010} and \cite{Specia_EAMT:2011}. The former consists of 4,000 English sentences from Europarl \cite{Koehn:05b}, which were translated into Spanish by a PBMT system and manually post-edited by professional  translators. The latter, which covers the news domain, includes 2,525  French--English  PBMT translations  and  1,000 English--Spanish translations with professional post-edits. 

Other useful data have been released by the organisers of the WMT APE task.
The first round of the task
\cite{bojar-EtAl:2015:WMT} presented participants with around 12,000 English--Spanish training data drawn from the news domain, with translations derived from a PBMT system. A peculiarity of this corpus is that post-edits were collected from a non-professional crowdsourced  workforce, with possible drops in terms of reliability and consistency.\footnote{This is a possible cause of the poor results achieved by participants: none of them was indeed able to beat the APE task baseline represented by a ``\textit{do-nothing}'' system that leaves all the raw MT translations unmodified.}

The second round of the task
\cite{bojar-EtAl:2016:WMT1} presented participants with a corpus released within the EU project QT21,\footnote{\url{http://www.qt21.eu/}} the same used for the WMT'16 quality estimation task. It comprises  13,000 English--German training data drawn from the information technology (IT) domain, with source sentences translated by a PBMT system and post-edits collected from professional translators. The combination of domain specificity and higher post-editing quality resulted in significant gains over the baseline.

The third round of the task
\cite{bojar-EtAl:2017:WMT1} focused on both English--German and German--English data (also in this case provided by the QT21 project \cite{MTsummit17} and shared with the WMT'17 quality estimation task). English--German training data are drawn from the IT domain and consist of around 13,000 triplets. German--English training data, instead, come from the medical domain and comprise around 26,000 triplets. In both cases, translations were produced by a customised PBMT system and post-edited by professional translators.

\subsection{Synthetic corpora}
The use of synthetic resources aims to overcome the aforementioned problem of ``gold'' data scarcity with approximate solutions. This can be done in different ways. Several previous works have shown the viability of mimicking the ideal scenario in which the training triplets include actual human post-edits of  machine-translated text by learning, instead, from the weaker connection between the MT output and external references. Though with variable margins, \cite{Oflazer:2007,Bechara2011,rubino2012statistical} report translation quality improvements in the PBMT scenario with post-editing components trained on  (\textit{source}, \textit{MT}, \textit{reference}) triplets. To the best of our knowledge, though potentially useful to APE research, none of such previous works released reusable datasets.

When moving to the data-demanding neural framework, data scarcity becomes a major problem that definitely calls for the external support of artificial  corpora that are orders of magnitude larger than the current training sets.

\paragraph{}
A widely used resource, described in \cite{junczysdowmunt-grundkiewicz:2016:WMT}, was included in the training set of the winning (and almost all) submissions to the last two English--German rounds of the APE task at WMT (IT domain). It consists of 4.3 million instances created by first filtering a subset of IT-related sentences from the German Common Crawl corpus\footnote{\url{commoncrawl.org}}, and then by using two  English--German and German--English  PBMT  systems trained on in-domain IT corpora for a round-trip translation of the selected sentences (De$\rightarrow$En$\rightarrow$De). The final triplets were created by using: \textit{i)} the English translations as (artificial) \textit{source} sentences, \textit{ii)} the round-trip  German translations as (artificial) uncorrected \textit{MT} output, and \textit{iii)} the  original  German sentences as (artificial) \textit{post-edits}.

By construction, this artificial data set approximates the quality of gold corpora by trying to keep a weak connection between the ``post-edits'' and the MT output. Keeping such connection, however, comes at the cost of having two levels of potential noise in the data, namely the possible  errors introduced by the German--English translation needed to produce the \textit{source} element of each triplet, and those of the English--German translations performed to produce the \textit{MT} output. The approach we adopted to create the eSCAPE corpus, instead, follows a different strategy. As described in the next section, we start from parallel data and perform one single automatic translation step to produce the \textit{MT} element of our triplets. The connection between MT output and ``post-edits'' is hence weaker than in \cite{junczysdowmunt-grundkiewicz:2016:WMT} due to the fact that our ``post-edits'' are actually independent reference translations of the source sentences. However, the possible noise introduced by translation errors can only affect one element of our triplets. Analysing the trade-off between translation noise and \textit{MT}-\textit{post-edits} proximity is out of the scope of this work but it is definitely an interesting aspect for future investigations.

\section{The eSCAPE Corpus}
\label{sec:escape}
The eSCAPE corpus\footnote{\url{http://hltshare.fbk.eu/QT21/eSCAPE.html}} consists in two datasets (En-De and En-It) made of (\textit{source}, \textit{MT}, \textit{reference}) triplets, where the \textit{MT} segment is obtained by translating the \textit{source} both with phrase-based and neural MT models.
Its creation started from parallel (\textit{source}, \textit{target}) data collected from the WEB by merging several corpora belonging to various domains. Table \ref{tab:eSCALE} lists all the corpora used, indicating their domain and size in terms of number of sentences. Since some data sets, such as PatTR and Common Crawl, are only available in one language pair (En-De), the total number of sentences is different between the two language directions (En-De is twice larger than En-It). Apart from PatTR\footnote{\url{http://www.cl.uni-heidelberg.de/statnlpgroup/pattr/}} and Common Crawl, all the datasets are available in the OPUS repository\footnote{\url{http://opus.lingfil.uu.se}}. 
Before building the translation systems and producing the \textit{MT} segments, all the corpora reported in Table \ref{tab:eSCALE} have been concatenated and shuffled (to avoid blocks of sentences belonging to the same domain) removing duplicates. This resulted in 7,258,533 English--German and 3,357,371 English--Italian sentence pairs.\\

\begin{table}[ht]
\begin{center}
\begin{small}
\begin{tabularx}{\columnwidth}{|l|X|X|X|}
\hline
\textbf{Corpus} & \textbf{Domain} & \textbf{En-De} & \textbf{En-It}\\
\hline
Europarl v7 & LEGAL& 1,920,209 & 1,909,115\\
\hline
ECB & LEGAL& 11,317 & 193,047\\
\hline
Common Crawl & MIXED & 2,399,123 & -\\
\hline
JRC Acquis & LEGAL& 719,372  & 810,979\\
\hline
News Commentary v11 & NEWS& 242,770 & 40,009\\
\hline
Ted Talks & MIXED & 143,836 & 181,874\\
\hline
EMEA & MEDIC. & 1,108,752 & 1,081,134 \\
\hline
PatTR in-domain & MEDIC. &  1,848,303 & -\\
\hline
Wikipedia Titles & MEDIC.  & 10,406 & - \\
\hline
Gnome & IT &28,439 & 319,141\\
\hline
Ubuntu & IT & 13,245 & 21,014\\
\hline
KDE4 & IT & 224,035 & 175,058\\
\hline
OpenOffice & IT & 42,391 & - \\
\hline
PHP & IT & 39,707 & 35,538 \\ 
\hline
\hline
TOTAL & & 8,853,762 & 4,128,128\\
\hline
\end{tabularx}
\caption{List of datasets merged in the eSCAPE corpus.}
\label{tab:eSCALE}
\end{small}
 \end{center}
\end{table}



\subsection{MT systems}
Driven by the need of translating these large quantities of source segments, the ModernMT toolkit \cite{bertoldi2017mmt} has been used as translation system to generate both phrase-based and neural outputs.
ModernMT is a new open-source MT software that consolidates the current state-of-the-art MT technology into a single and easy-to-use product. The toolkit adapts to the context in real-time and is capable of learning from  (and evolving through) interaction with users, with the final aim of increasing MT-output utility for the translator in a real professional environment. 

To avoid the risk of translating source segments that are in the training set, the collected sentence pairs were split in \textit{4} slices: \textit{3} parts were used to train the MMT models and the remaining one was translated. In this cross-validation setting, one sentence pair has been processed once for each experiment, either in training or in test.

For \textbf{phrase-based MT}, ModernMT uses high-performance embedded databases to store parallel and monolingual language data and associated statistics. Instead of pre-computing phrase-based feature function scores, these are computed on the fly, at translation time, from raw statistics. This allows the MMT toolkit to significantly speed up the training and test processes, to easily scale to large quantities of data, and to  adapt on-the-fly to new domains. 
Training and test of the phrase-based models were run in parallel on several CPUs for around one week. Final performance, computed on a subset of the data, is 36.76 BLEU points for English--German and 38.08 for English--Italian.

For \textbf{neural MT}, the toolkit builds on the extension of a generic neural MT system based on  the Nematus  toolkit \cite{sennrich-EtAl:2017:EACLDemo}\footnote{\url{https://github.com/rsennrich/nematus}}, implementing  the  encoder-decoder-attention model architecture by \cite{DBLP:journals/corr/BahdanauCB14}.
Such extension consists in an internal dynamic memory, storing external user translation memories (TMs). When ModernMT receives a translation query, it quickly analyses its context, recalls from its memory the most related translation examples, and instantly adapts its neural network to the query \cite{FARA-EACL17}.
Training and test of the neural models were run  on one GPU (NVIDIA Tesla K80) for around three weeks. Final performance is 38.17 BLEU points for English--German and 41.01 for English--Italian. 

To give the possibility for experiments on domain-adaptation for APE, each eSCAPE triplet is associated to a label indicating the name of the corpus from which the original (\textit{source}, \textit{reference}) pair was extracted.\\

\section{Experiments}
\label{sec:exp}
To test the usefulness of the eSCAPE corpus, we run APE experiments for
both the language pairs covered by the data set. 
En-De and En-It data were first tokenised and then split into dev (2,000 triplets), test (10,000) and training (the remaining instances).
For the sake of comparison,  we performed the same data splits for both the phrase-based and for the neural-based section of the corpus.
%

As APE system, we chose the best system at this year's round of the WMT APE shared task \cite{FBK:2017:WMT}. It consists in a neural multi-source model, in which the source and the MT segment are encoded separately and then merged together by a feed-forward network layer. A shared dropout is applied to both source and MT encoders. 
In this multi-source architecture both the encoders are trained jointly.

 In our experiments, the hyper-parameters of all the systems in both language directions were the same. 
The vocabulary was created by selecting 50,000 most frequent sub-words, following the BPE approach of \newcite{sennrich2016neural}. 
Word-embedding and GRU (gated recurrent unit) hidden-state sizes were both set to 1024.
Network parameters were optimized with Adagrad \cite{adagrad} with a learning rate of 0.01. Source and target dropout was set to 10\%, whereas encoder and decoder hidden states, weighted source context, and embedding  dropout was set to 20\% \cite{sennrich-haddow-birch:2016:WMT}. The batch size was set to 100 samples, with a maximum sentence length of 50 sub-words.
 
During training, the performance of the APE system was monitored on the development set and, at the end of the training phase,  the model with highest BLEU score was used to post-edit the test set.  The results are reported in 
Table 3 where,
for both the language pairs and for both phrase-based and neural-based artificial data,
the performance of the APE systems is compared against the ``\textit{do-nothing}'' APE baseline (i.e. a system that leaves all the raw MT output unmodified).  



\begin{table}[ht]
\centering
\label{tab:APE}
\begin{tabular}{l|l|l}
                    & En--De & En--It \\ \hline
\multicolumn{3}{c}{Phrase-based MT}   \\ \hline
Do-Nothing baseline & 36.76  & 38.08  \\
APE                 & \textbf{38.15}  & \textbf{39.80}  \\ \hline
\multicolumn{3}{c}{Neural MT}         \\ \hline
Do-Nothing baseline & 38.17  & 41.01  \\ 
APE                 & \textbf{39.21}  & \textbf{42.15} 
\end{tabular}
\caption{Neural APE  results (BLEU score improvements are statistically significant with $p<0.05$ computed with paired bootstrap resampling  \protect\cite{koehn2004statistical}).}
\end{table}





It is interesting to note that APE systems outperform the baselines in both language settings with  statistically significant gains.\footnote{Although we consider the measured gains as a good indicator of the usefulness of using eSCAPE for  training APE models, a study involving human evaluation would allow us to draw definite conclusions. Such a costly study, however, falls out of the scope of this paper and is left for future work.} This holds true both when they are trained and tested on artificial data built from phrase-based models (+1.39 on En--De, +1.72 on En--It), and when training and test are performed on artificial data derived from neural models (+1.04 on En--De, +1.14 on En--It).\footnote{Though interesting, other settings in which the two sections of eSCAPE are either combined or alternatively used one for training and one for test fall out of the scope of this paper and are left for future investigation.}

The observed gains  vary for the two language pairs (with highest results on En--It) and depending on the type of data used. Concerning this latter aspect, the higher quality of neural MT output results in lower gains on both language settings. This confirms previous outcomes from the WMT APE task: the higher the baseline (i.e. the BLEU score of the raw MT output), the lower the number of correction patterns that can be learned from the training data and the possibility to leverage their applicability to test data \cite{bojar-EtAl:2017:WMT1}.

Differently from the most recent shared evaluation settings (\textit{i.e.} WMT'16 and WMT'17), in which neural APE has been tested in narrow domains, our results indicate that APE systems trained on the eSCAPE corpus can be also effective in the more challenging mixed-domain condition, where the correction rules are sparse across different domains, hence difficult to be learned and generalized. Considering the negative outcomes of the WMT'15 pilot task, which proposed a challenging evaluation setting based on general news data in which none of the participants was able to beat the ``\textit{do-nothing}'' baseline, this is an interesting finding that adds value to our resource. 

The BLEU score improvements also confirm the findings of \cite{Oflazer:2007,Bechara2011,rubino2012statistical} and extend them to neural APE. In fact, they report translation quality improvements in the PBMT scenario with an APE trained on  \textit{source}, \textit{MT}, and independent \textit{reference}. This suggests that, despite the aforementioned weak connection between the MT output and the ``post-edits'' of our triplets, APE models can be effectively trained on the eSCAPE corpus.

\section{Conclusion and Future Work}
\label{sec:concl}
We presented the eSCAPE corpus, a  large-scale \textit{S}ynthetic \textit{C}orpus for \textit{A}utomatic \textit{P}ost-\textit{E}diting consisting of millions of (\textit{source}, \textit{MT}, \textit{post-edit}) triplets created via  machine translation. eSCAPE is designed to support the recent trends in automatic post-editing, which show a clear predominance of data-demanding neural approaches. To cope with such demand, the current version of the corpus  contains millions of triplets for two language pairs: English--German (14.4 millions)  and English--Italian (6.6 millions). For both language pairs, half of the artificial data is obtained via phrase-based translation, while the other half is produced by better performing neural MT models. 
Having the same source sentences translated with both paradigms aims to enable future comparisons in the application of APE technology on the two types of output.
The size of the corpus (the largest of its kind) is expected to ease model training and yield further state of the art improvements.
 Our preliminary experiments on mixed-domain data confirm this expectation: though trained on artificially-created instances, APE models significantly outperform baseline results in both language directions, independently from the MT technology underlying the data generation process. 
The work reported in this paper is the initial step of a more ambitious roadmap aimed to extend the resource with more data covering a larger spectrum of domains and language combinations. The current version of eSCAPE can be freely downloaded from: \url{http://hltshare.fbk.eu/QT21/eSCAPE.html}.

\section*{Acknowledgments}
This work has been partially supported by the EC-funded H2020 projects QT21 (grant  agreement  no. 645452) and  ModernMT  (grant  agreement  no. 645487).
The authors gratefully acknowledge the support of NVIDIA Corporation with the donation of the Tesla K80 GPU used for this research.

\section{Bibliographical References}
\label{main:ref}
\bibliographystyle{lrec}
\bibliography{xample}
\end{document}